\documentclass[preprint,12pt,authoryear]{elsarticle}

\usepackage{amssymb}
\usepackage{amsmath}
\usepackage{xcolor}
\usepackage{booktabs}
\usepackage{multirow}
\usepackage{graphicx}
\usepackage{hyperref}
\usepackage{float}

\journal{ARXIV}

\begin{document}

\begin{frontmatter}

\title{An Explainable LLM Agent Layer for Open-World Anomaly Detection in Oil Wells}

\author[1]{Lucas Gouveia Omena Lopes\corref{mycorrespondingauthor}}
\cortext[mycorrespondingauthor]{Corresponding author}
\ead{lucas.lopes@ctec.ufal.br}
\author[2]{Thales Miranda de Almeida Vieira}
\author[1]{Eduardo Toledo de Lima Junior}
\author[1]{William Wagner Matos Lira}

\affiliation[1]{organization={Laboratory of Scientific Computing and Visualization, Federal University of Alagoas},
            city={Maceio},
            state={Alagoas},
            country={Brazil}}
\affiliation[2]{organization={Computing Institute, Federal University of Alagoas},
            city={Maceio},
            state={Alagoas},
            country={Brazil}}

\begin{abstract}
Open-World Learning (OWL) pipelines for oil well anomaly detection have recently been shown to combine autoencoder-based detection, multiclass classification, and Mahalanobis-based novelty detection on the public 3W dataset \citep{Lopes2025EAAI, Lopes2026OTC}. These pipelines answer \textit{what happened}, but they do not explain \textit{why the model believes it} or \textit{what the operator should do next}, and they do not put a human-readable name on the novelty clusters they discover.

This paper evaluates a Large Language Model (LLM) agent layer placed downstream of the OWL pipeline, designed as a \textbf{companion} to the published upstream methods rather than a replacement. Using the Qwen3.5-397B-A17B Mixture-of-Experts model served via NVIDIA NIM, the agent receives structured sensor metrics and upstream classification or novelty assertions, and returns natural-language justifications, confidence-ranked critiques, and consolidated names for detected novelties. Across three studies spanning 989 real well-file segments from the 3W dataset, the agent achieved $35.1\%$ top-1 / $63.9\%$ top-3 (95\% CI [56.9, 70.4]) classification on all nine classes, $71.7\%$ top-2 validation [64.8, 77.6] with precision $0.91$ [0.84, 0.95] across 7 probed classes, and $89.7\%$ novelty detection [87.0, 91.9] with stable cluster naming on 5 of 7 hidden classes.

The agent is not a standalone classifier. Its role is to: (1) confirm upstream decisions when sensor evidence supports them, (2) justify decisions in sensor-grounded language operators can audit, (3) flag disagreement when upstream labels are implausible, and (4) name novelties so that clustered unlabeled events arrive at the engineer with a consolidated human-readable label. The goal is to close the explainability gap that currently blocks deployment of OWL pipelines in operational settings.
\end{abstract}


\end{frontmatter}

%% ============================================================
\section{Introduction}\label{sec:intro}
%% ============================================================

The oil and gas industry relies on continuous real-time monitoring of sensor data---temperature, pressure, and flow rates---to maintain safe and efficient operations \citep{VARGAS2019, huffner2019}. Anomalies in oil well production can indicate faults, inefficiencies, or safety risks with potentially severe consequences \citep{GUILHERME2011201}. Detecting, classifying, and responding to these anomalies is therefore critical.

Machine learning methods have been widely applied to anomaly detection in oil wells. The 3W dataset, introduced by \citet{VARGAS2019} and recently updated to version 2.0 \citep{Vargas2025_3w2}, is the largest labeled open dataset of anomalies in oil well production and has supported extensive research in anomaly detection, predictive maintenance, and multivariate time-series classification \citep{marins2020fault, Carvalho2021flow, Aranha2024}. However, most existing methods focus on what anomaly occurred, without providing explanations of \textit{why} the model reached its conclusion or \textit{what} the operator should do in response.

\citet{Lopes2025EAAI} introduced the first Open-World Learning (OWL) strategy for oil well anomaly detection, integrating autoencoder reconstruction error, binary classifiers, and clustering methods to both classify known anomalies and discover novel ones---achieving 81\% global clustering accuracy with up to 99\% on updated binary classifiers. \citet{Lopes2026OTC} advanced this foundation significantly with a 1D U-Net segmentation approach and, critically, a hybrid Mahalanobis novelty detector operating on the latent space of a multiclass classifier's penultimate layer, making novelty detection substantially more reliable. Together, these works establish a complete pipeline from detection through classification to novelty grouping.

Despite these advances, both works identify remaining challenges: binary classifiers remain overconfident on unseen events \citep{Lopes2025EAAI}, the Mahalanobis detector provides only a numerical distance without rationale \citep{Lopes2026OTC}, grouped novelty clusters are unlabeled \citep{Lopes2025EAAI, Lopes2026OTC}, and operator-driven validation limits scalability \citep{Lopes2025EAAI}. This paper addresses these gaps by evaluating a Large Language Model (LLM) agent layer that sits downstream of the upstream OWL pipeline and provides the explainability, validation, and naming functions the pipeline cannot produce on its own.

\subsection{LLMs in oil and gas and anomaly detection}\label{sec:llm_related}

The application of LLMs to the oil and gas domain is an emerging research area. Recent work includes LLM-based agents for natural gas leakage detection \citep{Wei2024leakage}, information extraction from historical well records \citep{Ma2024wellrecords}, intelligent data analysis for production optimization \citep{Liu2024oilgasLLM}, and domain-specialized models such as EnergyGPT \citep{Chebbi2025energygpt}.

In the broader anomaly detection literature, LLMs have been applied to time-series anomaly detection with explainability as a primary goal. Adaptive and explainable AI agents using LLM-enhanced contextual reasoning have been proposed for critical IoT infrastructure \citep{Sharma2025xaiagents}. Agentic and multi-agent architectures for multimodal anomaly detection represent a growing research direction \citep{Belay2026agenticAD}. LLM-assisted logic rule learning has been explored for encoding human expertise into interpretable rules \citep{Zhang2026llmrules}, and human-in-the-loop LLM frameworks have shown effectiveness for industrial time-series fault diagnosis \citep{Zhang2025LLMTSFD}. The classical novelty-detection methods these works build on are surveyed by \citet{Pimentel2014review}.

However, the specific application of LLMs as \textbf{companion explainability layers} for existing ML pipelines in oil well monitoring---where the LLM does not replace the upstream classifier but rather justifies, validates, and names its outputs---has not been previously explored. This work bridges that gap.

\subsection{Contributions}\label{sec:contributions}

The main contributions of this paper are:

\begin{enumerate}
    \item A method for computing structured sensor metrics from preprocessed 3W data and building data-driven class profiles that serve as the LLM's knowledge base.
    \item A systematic evaluation across three studies (classification, validation, and novelty detection) of a general-purpose LLM (Qwen 3.5 MoE) as a companion to an established OWL pipeline.
    \item Evidence that the LLM can provide sensor-grounded justifications, validate or reject upstream classifications with 91\% precision, and generate consolidated names for novelty clusters at 89.7\% detection rate.
    \item Analysis of the LLM's failure modes on physically ambiguous classes (4, 5, 9), demonstrating that these failures are physics-driven rather than model-driven.
\end{enumerate}

%% ============================================================
\section{Delimitations}\label{sec:delimitations}
%% ============================================================

This study evaluates the LLM agent layer using only the public 3W dataset \citep{VARGAS2019}. The proprietary ICV Fault data used in \citet{Lopes2025EAAI} is not included in this evaluation; extending to proprietary anomaly types is straightforward but has not been done here.

The LLM (Qwen 3.5 MoE 397B) is used as a general-purpose model without fine-tuning on oil well data. The agent receives only structured metrics (not raw time series) and textual class profiles. It does not have access to spatial features, well geometry, or operational context beyond the four sensors (P-PDG, P-TPT, T-PDG, T-TPT).

The evaluation measures the LLM as a companion to the upstream OWL pipeline, not as a standalone classifier. The per-class Study~0 numbers make this delimitation explicit: on classes where the sensor signature is ambiguous (notably class~9 with only 7\% top-1), the LLM cannot reliably classify alone.

%% ============================================================
\section{Proposed Method}\label{sec:method}
%% ============================================================

\subsection{Dataset and preprocessing}\label{sec:data}

The evaluation uses the public 3W dataset \citep{VARGAS2019} with the same four sensors and preprocessing as the upstream papers: z-score normalization followed by min-max scaling to $[0, 1]$, wavelet denoising (Daubechies db4, level~3, soft thresholding), and missing-value handling (forward fill, backward fill, zero fill for remaining NaN). Segments are drawn primarily from real WELL files. For classes with fewer than 30 real instances, simulated data from the 3W dataset were used to complement the real data, ensuring a minimum sample size for robust profile building and evaluation. Hand-drawn instances are excluded.

\subsection{Metric computation}\label{sec:metrics}

A critical design decision is that the LLM receives \textbf{structured numerical metrics}, not raw time series. For each event segment, a shared metrics module computes per-sensor and cross-sensor features from the preprocessed data, ensuring exact numerical parity between knowledge building and evaluation.

\subsubsection{Per-sensor metrics}

Given a preprocessed event segment with baseline window $\mathbf{b}$ (timesteps before event onset) and event window $\mathbf{e}$ (timesteps during the event), the following metrics are computed for each of the four sensors:

\begin{equation}
\delta = \bar{e} - \bar{b}, \qquad
\text{pct} = \frac{\delta}{|\bar{b}|} \times 100, \qquad
z = \frac{|\delta|}{\sigma_b}
\end{equation}

\noindent where $\sigma_b = \max(\text{std}(\mathbf{b}), 10^{-6})$ guards against zero-variance baselines. The slope is computed via least-squares linear fit over the event timesteps.

Categorical labels are derived from these quantities using fixed thresholds:

\begin{itemize}
    \item \textbf{Direction}: \texttt{increase} if $\delta > 0.5\sigma_b$, \texttt{decrease} if $\delta < -0.5\sigma_b$, else \texttt{stable}.
    \item \textbf{Magnitude}: \texttt{small} if $z < 1$, \texttt{moderate} if $z < 3$, else \texttt{large}.
    \item \textbf{Rate}: \texttt{sudden} if $|\text{slope}| > 0.01$, \texttt{fast} if $> 0.002$, else \texttt{gradual}.
    \item \textbf{Behavior}: composite label from oscillation ratio, slope, and step-change analysis.
\end{itemize}

Additional metrics include the oscillation ratio (fraction of sign changes in the first difference of $\mathbf{e}$) and noise ratio ($\sigma_e / \sigma_b$). The complete set of 13 per-sensor metrics is summarized in Table~\ref{tab:metrics}; cross-sensor metrics are described in the following subsection.

\begin{table}[ht!]
\centering
\caption{Per-sensor metrics computed for each event segment.}
\label{tab:metrics}
\begin{tabular}{ll}
\toprule
\textbf{Metric} & \textbf{Description} \\
\midrule
\texttt{delta\_from\_baseline} & $\bar{e} - \bar{b}$ (normalized units) \\
\texttt{pct\_variation} & Percentage change relative to $\bar{b}$ \\
\texttt{z\_score} & $|\delta| / \sigma_b$ \\
\texttt{slope} & Least-squares linear slope over event window \\
\texttt{direction} & Categorical: increase / decrease / stable \\
\texttt{magnitude} & Categorical: small / moderate / large \\
\texttt{rate} & Categorical: sudden / fast / gradual \\
\texttt{behavior} & Composite: step\_change, sharp\_rise, etc. \\
\texttt{oscillation\_ratio} & Fraction of sign changes in $\Delta\mathbf{e}$ \\
\texttt{noise\_ratio} & $\sigma_e / \sigma_b$ \\
\texttt{event\_range} & $\max(\mathbf{e}) - \min(\mathbf{e})$ \\
\texttt{baseline\_mean} & $\bar{b}$ \\
\texttt{event\_mean} & $\bar{e}$ \\
\bottomrule
\end{tabular}
\end{table}

\subsubsection{Cross-sensor metrics}

The pressure relationship between P-PDG and P-TPT is classified as \texttt{same\_direction}, \texttt{diverging}, or \texttt{one\_stable\_or\_both\_stable} based on their individual direction labels. The strongest sensor (largest $|\delta|$) is also identified.

\subsection{Knowledge base (class profiles)}\label{sec:knowledge}

For each of the nine known anomaly classes, a knowledge profile is built by aggregating per-sensor metrics across all available training segments. The aggregation applies two-pass IQR-based outlier removal ($k=3.0$ then $k=1.5$) before computing summary statistics (mean, std, p25, p75, min, max) for numeric metrics and mode for categorical labels.

Each profile additionally includes:
\begin{itemize}
    \item \textbf{Macro behavior} per sensor: change type, noise level, consistency across files, dominant pattern.
    \item \textbf{Local behavior} per sensor: event shape, onset sharpness, recovery pattern, peak behavior.
    \item \textbf{General domain knowledge}: physical causes, risk factors, detection methods, prevention strategies, severity level, related anomalies, and industry context.
    \item \textbf{Distinguishing features}: auto-generated discriminative characteristics.
\end{itemize}

The full profile is serialized as JSON and rendered as a structured text prompt for the LLM. A sample profile (Class~2: Spurious Closure of DHSV) is shown in Appendix~\ref{app:profile}.

\subsection{LLM agent}\label{sec:agent}

\begin{itemize}
    \item \textbf{Model:} Qwen3.5-397B-A17B (Mixture-of-Experts; $\sim$397B total / $\sim$17B active parameters per token) \citep{Qwen2025} with thinking-mode enabled, served via the public NVIDIA NIM endpoint \texttt{qwen/qwen3.5-397b-a17b} (OpenAI-compatible API). Inference parameters: temperature $0.2$, top-p $0.7$, max output tokens $4096$. Thinking-mode chain-of-thought is consumed internally by the endpoint and is not used by downstream parsing; only the final JSON object is retained.
    \item \textbf{Input:} For each segment, the single-segment metrics (Section~\ref{sec:metrics}) alongside all visible class profiles (Section~\ref{sec:knowledge}).
    \item \textbf{Prompt:} The agent receives observed metrics, known class profiles with per-sensor statistics and distinguishing features, and task-specific instructions requesting ranked JSON output with justifications.
    \item \textbf{Output:} JSON with top-3 ranked candidates, per-candidate justification citing sensor evidence, confidence label (High/Medium/Low), and for novelty cases a \texttt{novel\_name} plus \texttt{novel\_description}.
\end{itemize}

\subsection{Study design}\label{sec:studies}

Three studies evaluate the agent's capabilities across complementary dimensions:

\begin{table}[ht!]
\centering
\caption{Study design overview.}
\label{tab:studies}
\begin{tabular}{clll}
\toprule
\textbf{Study} & \textbf{Scope} & \textbf{Segments} & \textbf{What is tested} \\
\midrule
0 & All 9 classes & 191 & Independent top-3 classification from metrics alone \\
1 & Classes 1--3, 6--9 & 187 & Validation of upstream proposals (confirm/reject) \\
2 & 15 pair holdouts + class 9 & 611 & Novelty detection and naming \\
\bottomrule
\end{tabular}
\end{table}

Classes 1, 2, 3, 6, 7, 8 form the ``active'' set; class~9 (\textit{Hydrate in Service Line}) is included as the canonical hard case the prior papers flag as a near-duplicate of class~8 \citep{Lopes2026OTC}. Classes 4 and 5 (\textit{Flow Instability} and \textit{Rapid Productivity Loss}) are excluded from Studies 1 and 2 because both prior works identify them as symptomatic rather than causal---their top-3 ambiguity in Study~0 is the clearest illustration of why. Study~0 retains all nine classes so the inherited symptomatic ambiguity can be quantified.

\begin{figure}[ht!]
\centering
\includegraphics[width=0.95\textwidth]{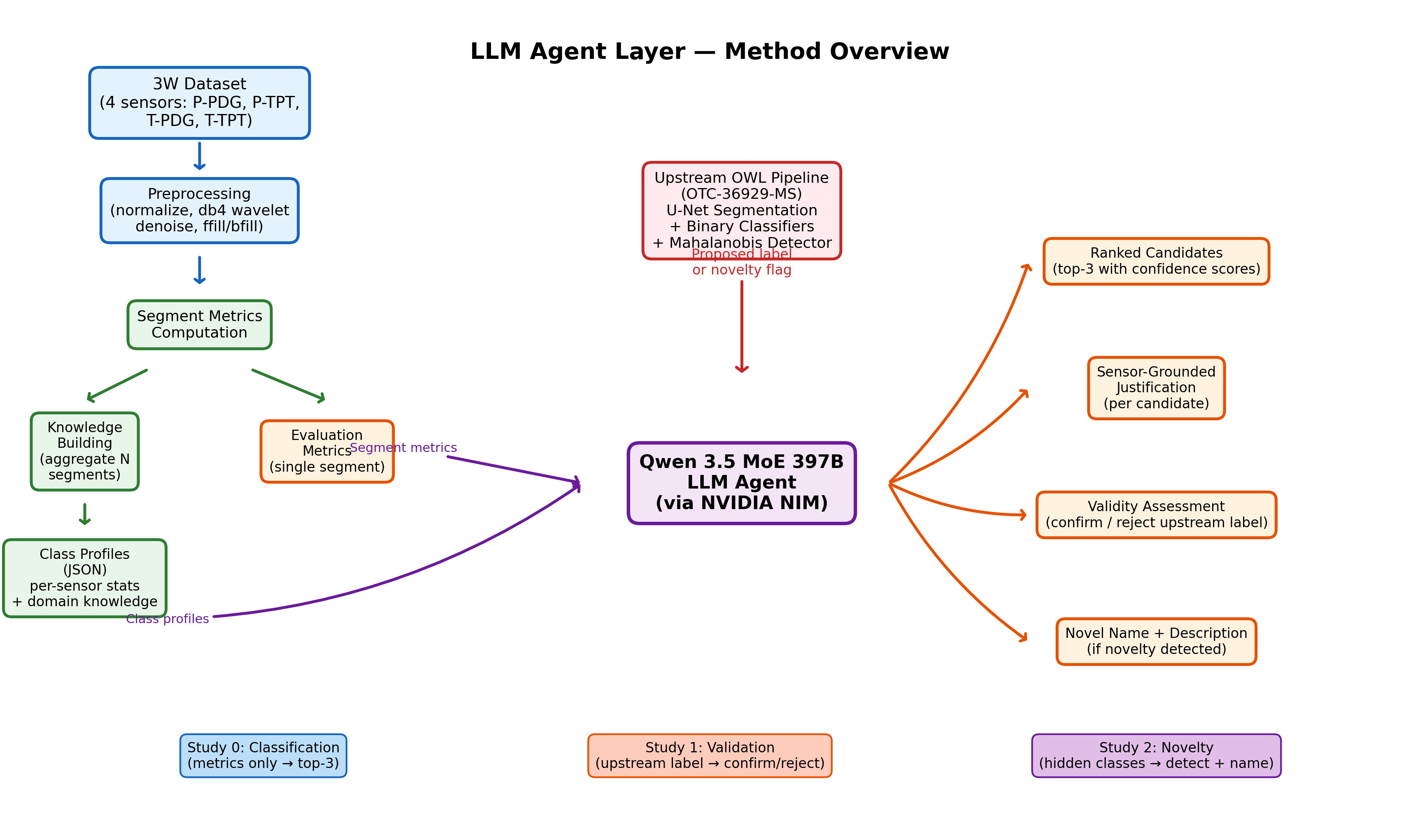}
\caption{Overview of the LLM agent layer. The agent receives structured metrics from the preprocessing pipeline and class profiles from the knowledge base, alongside upstream OWL pipeline outputs, and produces ranked candidates, justifications, validity assessments, and novelty names.}
\label{fig:method}
\end{figure}

%% ============================================================
\section{Results and Discussion}\label{sec:results}
%% ============================================================

\subsection{Study 0 --- Top-3 classification on all classes}\label{sec:study0}

\textbf{Headline:} 191 segments, top-1 = \textbf{35.1\%}, top-2 = \textbf{53.4\%}, top-3 = \textbf{63.9\%}.

\begin{table}[ht!]
\centering
\caption{Study 0 --- Top-k classification accuracy per class with Wilson 95\% confidence intervals (CI). Overall (all 9 classes, $n=191$): top-1 $35.1\%$ [28.7, 42.1], top-2 $53.4\%$ [46.3, 60.3], top-3 $63.9\%$ [56.9, 70.4].}
\label{tab:study0}
\small
\begin{tabular}{lrlll}
\toprule
\textbf{Class} & \textbf{n} & \textbf{Top-1 [95\% CI]} & \textbf{Top-2 [95\% CI]} & \textbf{Top-3 [95\% CI]} \\
\midrule
1 Abrupt Increase of BSW & 15 & 26.7\% [10.9, 52.0] & 40.0\% [19.8, 64.3] & 53.3\% [30.1, 75.2] \\
2 Spurious Closure of DHSV & 22 & \textbf{86.4\%} [66.7, 95.3] & 90.9\% [72.2, 97.5] & 100.0\% [85.1, 100] \\
3 Severe Slugging & 32 & 18.8\% [8.9, 35.3] & 53.1\% [36.4, 69.1] & 53.1\% [36.4, 69.1] \\
4 Flow Instability & 27 & 18.5\% [8.2, 36.7] & 25.9\% [13.2, 44.7] & 37.0\% [21.5, 55.8] \\
5 Rapid Productivity Loss & 15 & 26.7\% [10.9, 52.0] & 53.3\% [30.1, 75.2] & 66.7\% [41.7, 84.8] \\
6 Quick Restriction in PCK & 15 & 66.7\% [41.7, 84.8] & 66.7\% [41.7, 84.8] & 66.7\% [41.7, 84.8] \\
7 Scaling in PCK & 36 & 38.9\% [24.8, 55.1] & 66.7\% [50.3, 79.8] & 83.3\% [68.1, 92.1] \\
8 Hydrate in Production Line & 15 & 26.7\% [10.9, 52.0] & 53.3\% [30.1, 75.2] & 60.0\% [35.7, 80.2] \\
9 Hydrate in Service Line & 14 & 7.1\% [1.3, 31.5] & 14.3\% [4.0, 39.9] & 42.9\% [21.4, 67.4] \\
\bottomrule
\end{tabular}
\end{table}

The wide per-class CIs (e.g., class~9 top-1 spans 1.3--31.5\%) reflect the small per-class sample sizes inherent to the public 3W subset; differences across classes should be read as ordinal rather than precise point estimates.

\begin{figure}[ht!]
\centering
\includegraphics[width=0.95\textwidth]{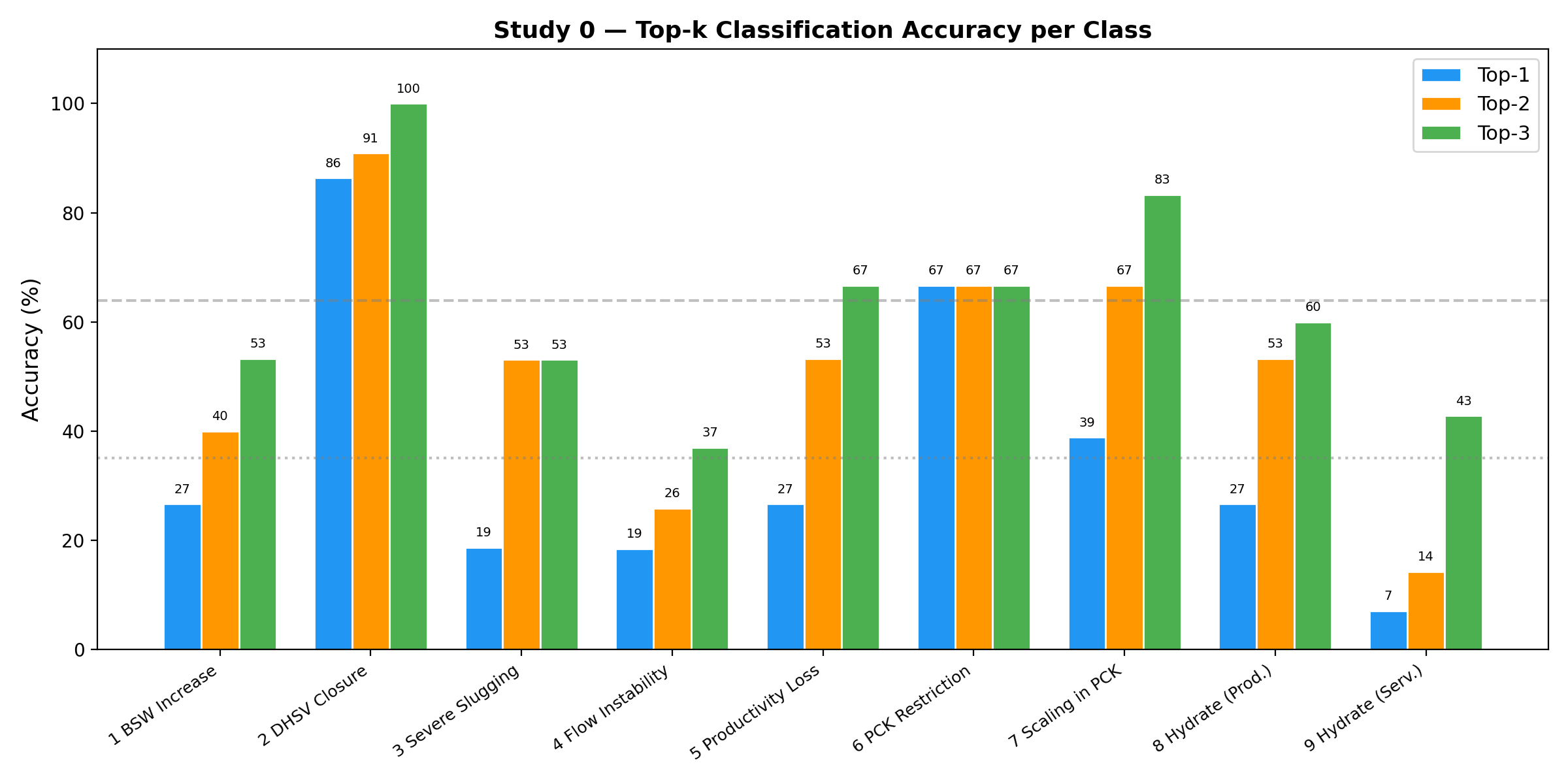}
\caption{Top-k classification accuracy per class in Study~0. Classes with clean sensor signatures (2, 6, 7) show strong performance; symptomatic classes (4, 5) and heterogeneous class~9 are weakest.}
\label{fig:topk}
\end{figure}

\begin{figure}[ht!]
\centering
\includegraphics[width=0.85\textwidth]{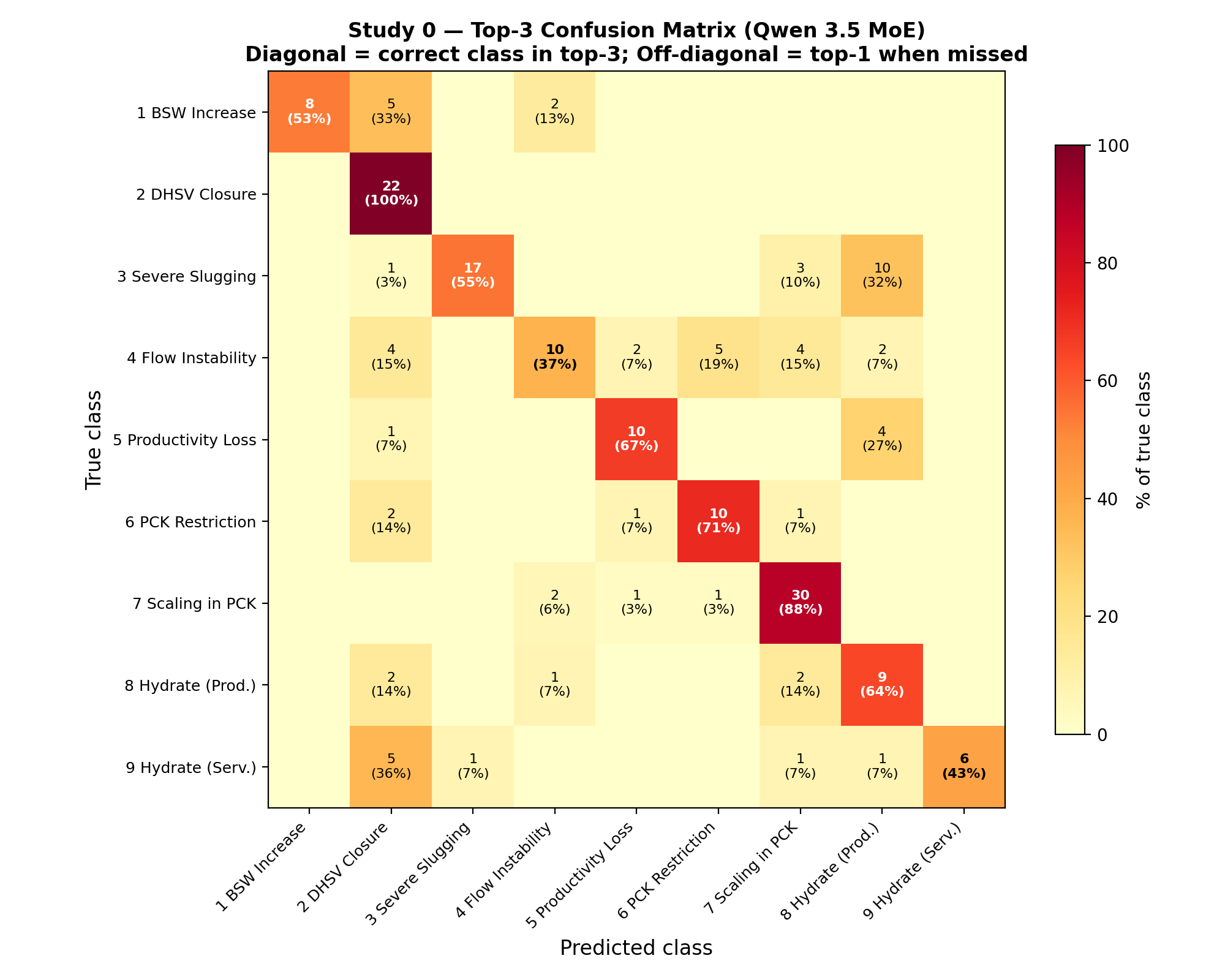}
\caption{Top-3 confusion matrix for Study~0. Each cell shows how many times a class appeared anywhere in the top-3 predictions and the percentage of top-3 slots it occupies per true class. Off-diagonal patterns reveal physically-motivated confusions---e.g., class~5 (Productivity Loss) absorbs predictions from class~3 (Slugging) and class~4 (Flow Instability) scatters across multiple classes.}
\label{fig:confusion}
\end{figure}

\begin{figure}[ht!]
\centering
\includegraphics[width=0.95\textwidth]{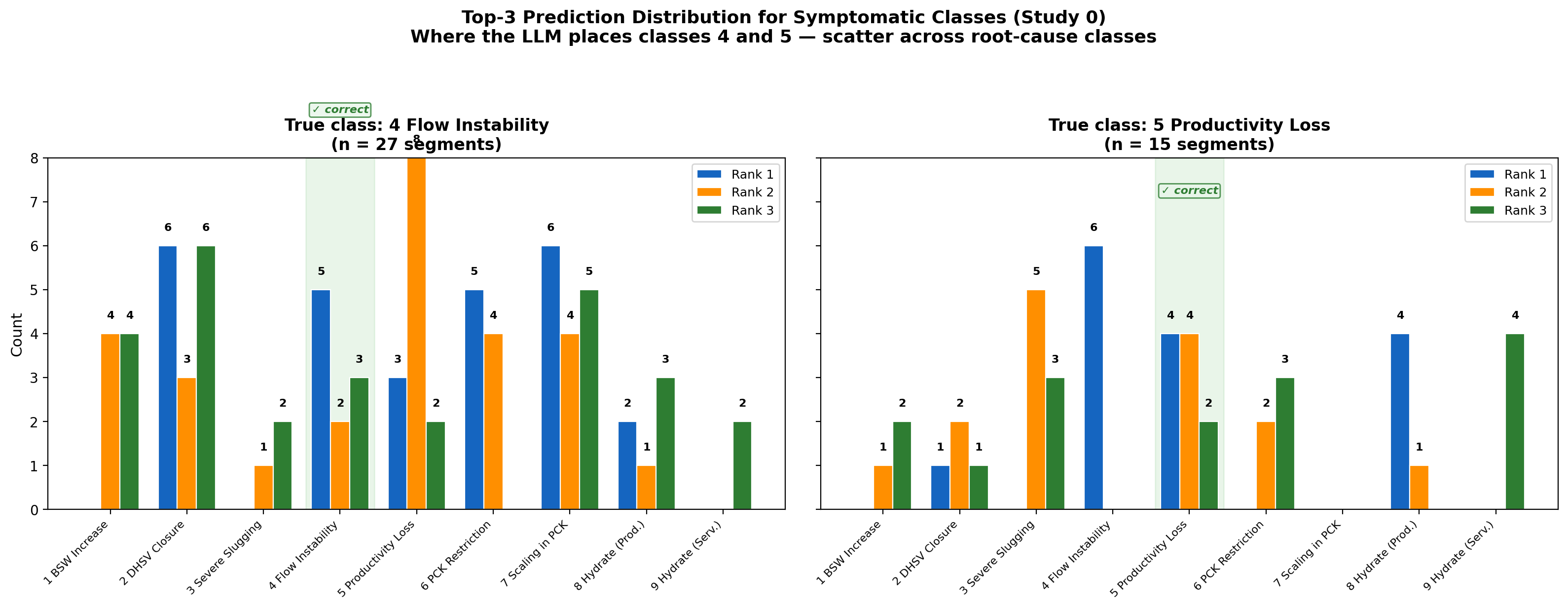}
\caption{Top-3 prediction distribution for symptomatic classes~4 and 5. Each bar shows how many segments placed a given class at rank~1, 2, or 3. The green-highlighted column marks the correct class. For class~4 (\textit{Flow Instability}), predictions scatter across classes~2 (DHSV), 5 (Productivity Loss), 6 (PCK), and 7 (Scaling)---all plausible root causes. For class~5 (\textit{Rapid Productivity Loss}), the agent frequently selects class~4 (Flow Instability) and class~3 (Severe Slugging) as alternatives, reflecting the physical overlap between symptom and cause.}
\label{fig:class45}
\end{figure}

\subsubsection{Why classes 4 and 5 drag the metric down}

Classes 4 (\textit{Flow Instability}) and 5 (\textit{Rapid Productivity Loss}) are the two worst-behaved classes in both upstream papers---\citet{Lopes2025EAAI} reports near-zero novelty detection accuracy on class~5, a problem only partially rescued by the Mahalanobis latent-layer detector introduced in \citet{Lopes2026OTC}. The agent inherits the same fundamental problem: these two classes are \textbf{symptoms, not root causes}.

\textit{Rapid Productivity Loss} is whatever emerges when the well stops producing. A DHSV closure (class~2), a hydrate plug (class~8 or 9), severe slugging (class~3), or a scaling event (class~7)---all eventually manifest as productivity loss. The sensor signature therefore overlaps with every one of them. \textit{Flow Instability} has the same problem: oscillations can be caused by slugging, partial PCK restriction, incipient scaling, or spurious valve operation.

The confusion matrix (Figure~\ref{fig:confusion}) and the per-rank breakdown (Figure~\ref{fig:class45}) show this plainly. For class~4, predictions scatter almost uniformly across classes~2, 5, 6, and 7; for class~5, classes~3 and 4 dominate the alternatives. In both cases the LLM is identifying the symptomatic layer correctly while the ground-truth labels point at the causal layer. This is why the upstream papers required dedicated Mahalanobis-latent rescue for class~5, and why this work excludes classes~4, 5, and 9 from the validation studies.

\textbf{Operational implication.} When a segment is classified as class~4 or 5, the agent should always be prompted for its top-3 root-cause candidates, which should be treated as a differential diagnosis rather than a definitive label.

\subsection{Study 1 --- Top-2 validation on classes 1--3, 6--9}\label{sec:study1}

\textbf{Headline:} 187 cases (149 correct + 38 wrong proposals) across 7 probed classes. The 38 wrong proposals were generated by, for each segment, sampling uniformly at random one of the remaining six probed classes (i.e., any class different from the ground-truth label). This sampling protocol guarantees that the validity flag is exercised against arbitrary plausible-but-incorrect siblings rather than only the most confusable neighbour, providing a conservative estimate of rejection capability.

\begin{table}[ht!]
\centering
\caption{Study 1 --- Overall validation metrics with Wilson 95\% confidence intervals.}
\label{tab:study1_overall}
\begin{tabular}{lll}
\toprule
\textbf{Metric} & \textbf{Value (k/n)} & \textbf{95\% CI} \\
\midrule
Top-2 accuracy & \textbf{71.7\%} (134/187) & [64.8, 77.6] \\
Top-2 confirm (correct kept in top-2) & 71.8\% (107/149) & [64.1, 78.4] \\
Top-2 reject (wrong pushed out of top-2) & 71.1\% (27/38) & [55.2, 83.0] \\
Correct proposals judged VALID & 67.1\% (100/149) & [59.2, 74.1] \\
Wrong proposals judged INVALID & 73.7\% (28/38) & [58.0, 85.0] \\
Precision (validity flag) & 0.909 (100/110) & [0.841, 0.950] \\
Recall (validity flag) & 0.671 (100/149) & [0.592, 0.741] \\
F1 (validity flag) & \textbf{0.772} & --- \\
\bottomrule
\end{tabular}
\end{table}

Precision is computed over the validity flag: $\text{TP}=100$ correct proposals judged VALID, $\text{FP}=10$ wrong proposals erroneously judged VALID (i.e., $38-28$). The previously reported F1 of $0.80$ used a rounded precision of $0.91$; the recomputed F1 with the exact $0.909$ precision is $0.772$.

The agent produces two independent judgments in a single call: (1) top-2 ranking and (2) validity flag. The validity column prevents discarding a correct upstream classification when the LLM happens to rank a sibling higher. In deployment: \textbf{IN TOP-2 + VALID = high confidence confirmation; OUT OF TOP-2 + INVALID = high confidence rejection}; any other combination requires the engineer to read the justification.

\begin{table}[ht!]
\centering
\caption{Study 1 --- Per-class validation performance.}
\label{tab:study1_perclass}
\begin{tabular}{lcccccc}
\toprule
\textbf{Class} & \textbf{Top-1} & \textbf{Top-2} & \textbf{Wrong} & \textbf{ValOK} & \textbf{ValRej} \\
 & \textbf{confirm} & \textbf{confirm} & \textbf{rejected} & & \\
\midrule
1 BSW & 40\% & 53\% & 60\% & 53\% & 60\% \\
2 DHSV & \textbf{95\%} & \textbf{95\%} & 80\% & \textbf{95\%} & 80\% \\
3 Slugging & 62\% & 62\% & 40\% & 59\% & 60\% \\
6 PCK Restr. & 73\% & 73\% & 80\% & 73\% & 80\% \\
7 Scaling & 86\% & 89\% & \textbf{100\%} & 81\% & \textbf{100\%} \\
8 Hydrate (P) & 47\% & 60\% & 40\% & 53\% & 60\% \\
9 Hydrate (S) & 29\% & 43\% & \textbf{88\%} & 29\% & 75\% \\
\bottomrule
\end{tabular}
\end{table}

\begin{figure}[ht!]
\centering
\includegraphics[width=0.95\textwidth]{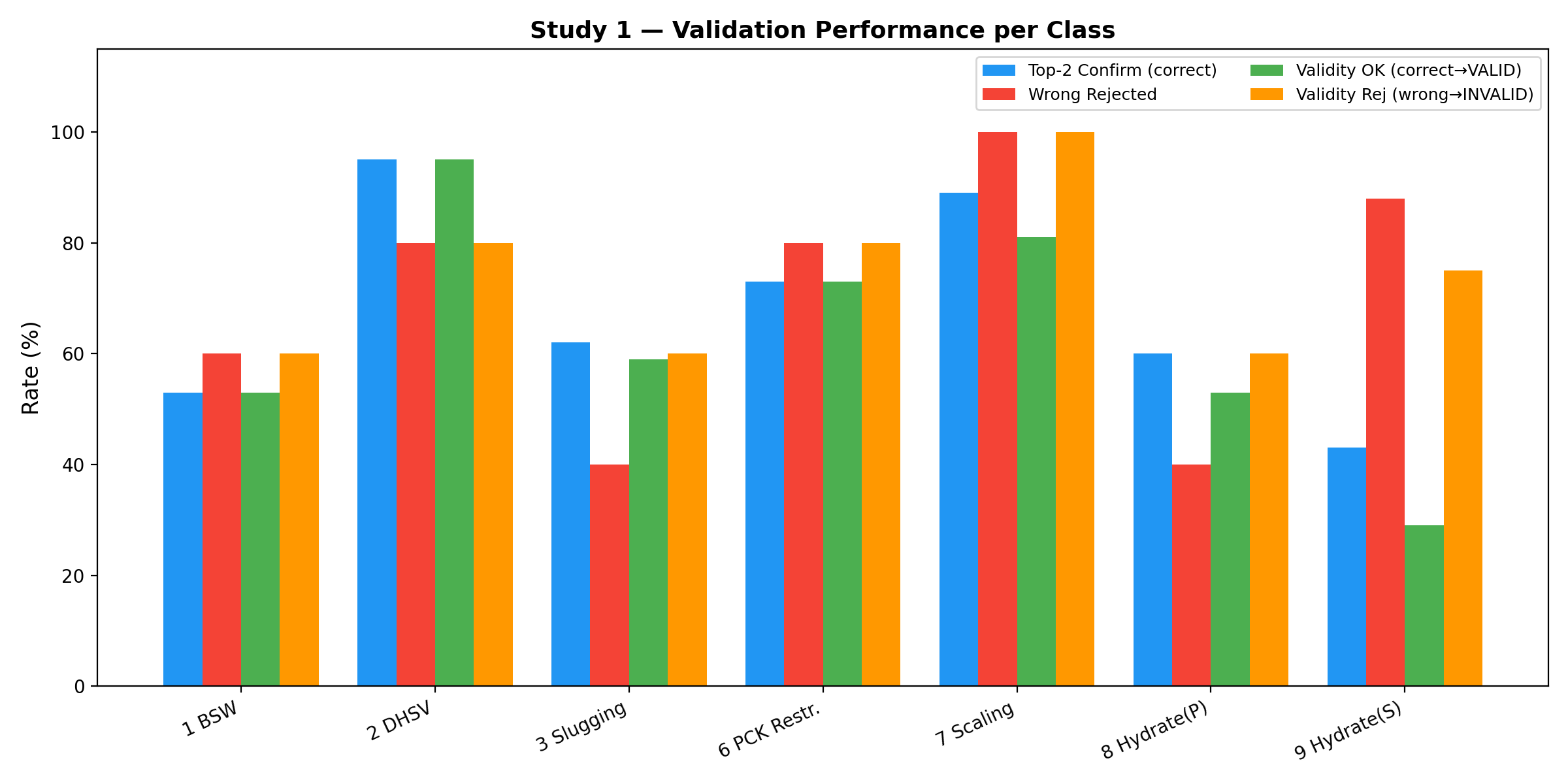}
\caption{Study 1 per-class validation. Clean-signature classes (2, 7) show near-perfect rates. Class~9 shows asymmetric behavior: low retention but excellent rejection.}
\label{fig:study1}
\end{figure}

\subsubsection{Class 9: a skeptic, not an affirmer}

Class~9 is the most informative row. Two patterns stand out:

\begin{itemize}
    \item \textbf{Low retention of correct proposals (43\%).} Real class-9 segments are extremely heterogeneous---including sensor-offline traces, P-PDG increases (plug above gauge), T-TPT spikes of +246\% (Joule--Thomson heating). The profile captures only the canonical ``all-sensors decrease'' archetype.
    \item \textbf{Excellent rejection of wrong proposals (88\%).} When a non-class-9 anomaly is proposed as class~9, the agent rejects it cleanly, including clean rejection of the near-duplicate class~8 case flagged by \citet{Lopes2026OTC}.
\end{itemize}

This asymmetry makes the agent a \textbf{skeptic} on class~9: it confirms only canonical-signature proposals and rejects atypical ones with defensible sensor-grounded rationale. In deployment, this is the right failure mode for a companion to an authoritative upstream classifier.

\subsection{Study 2 --- Novelty detection on classes 1--3, 6--9}\label{sec:study2}

\textbf{Headline:} 611 segments. Overall detection = \textbf{89.7\%} (548/611).

\begin{table}[ht!]
\centering
\caption{Study 2 --- Novelty detection rate per hidden class with Wilson 95\% confidence intervals.}
\label{tab:study2}
\begin{tabular}{lrll}
\toprule
\textbf{Class} & \textbf{n} & \textbf{Detection (k/n)} & \textbf{95\% CI} \\
\midrule
1 Abrupt Increase of BSW & 72 & 70.8\% (51/72) & [59.5, 80.1] \\
2 Spurious Closure of DHSV & 99 & \textbf{100.0\%} (99/99) & [96.3, 100.0] \\
3 Severe Slugging & 123 & 83.7\% (103/123) & [76.2, 89.2] \\
6 Quick Restriction in PCK & 70 & 95.7\% (67/70) & [88.1, 98.5] \\
7 Scaling in PCK & 175 & 94.9\% (166/175) & [90.5, 97.3] \\
8 Hydrate in Production Line & 61 & 91.8\% (56/61) & [82.2, 96.4] \\
9 Hydrate in Service Line & 11 & 54.5\% (6/11) & [28.0, 78.7] \\
\midrule
\textbf{Overall} & \textbf{611} & \textbf{89.7\%} (548/611) & \textbf{[87.0, 91.9]} \\
\bottomrule
\end{tabular}
\end{table}

The class~9 detection rate (54.5\%) carries a wide CI of [28.0, 78.7] due to the small sample ($n=11$) and should be interpreted as suggestive rather than conclusive.

\begin{figure}[ht!]
\centering
\includegraphics[width=0.85\textwidth]{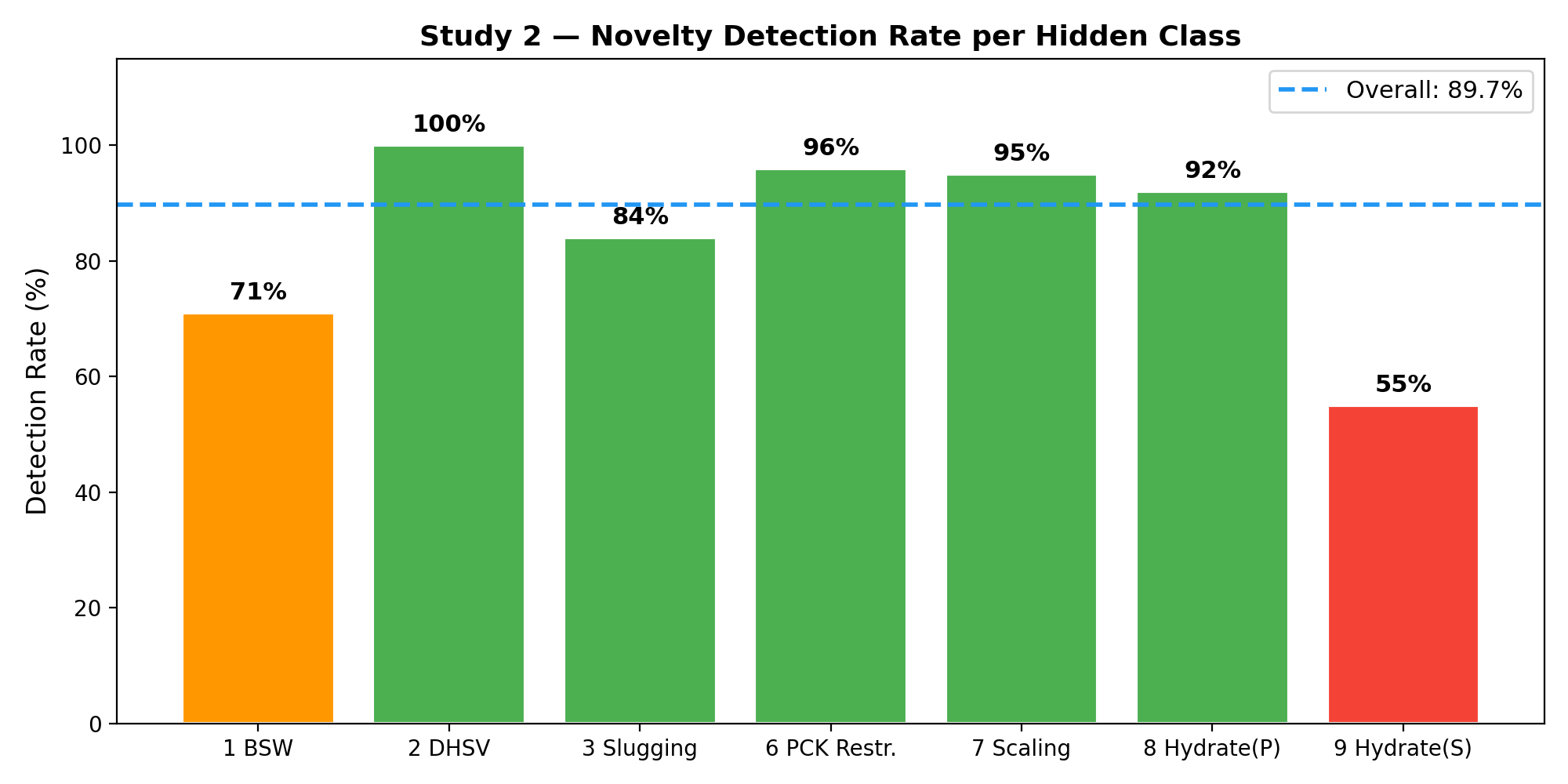}
\caption{Novelty detection rate per hidden class. Six of seven classes exceed 70\%; class~2 reaches 100\%.}
\label{fig:novelty}
\end{figure}

\subsubsection{Consolidated novelty names}

For each hidden class, we examined the \texttt{novel\_name} assigned across correct detections. A consolidated label emerged naturally for five of the seven classes. For class~1, all 51 names are unique---however, word-frequency analysis of the name corpus reveals dominant terms (\textit{thermal}: 16, \textit{pressure}: 13, \textit{wellhead}: 10, \textit{surge}: 10), enabling a cloud-derived consolidated label. Class~9 produced 6 distinct physically-grounded names, none dominant.

\begin{table}[ht!]
\centering
\caption{Consolidated agent-given names per hidden class.}
\label{tab:names}
\begin{tabular}{lp{7cm}}
\toprule
\textbf{True class} & \textbf{Consolidated name} \\
\midrule
1 BSW & Wellhead Thermal-Pressure Surge\textsuperscript{$\dagger$} \\
2 DHSV & Downhole Isolation Event with Wellhead Depressurization \\
3 Slugging & Production Rate Surge with Cyclic Pressure Decline \\
6 PCK Restr. & Downhole Telemetry Loss with Dual Pressure Surge \\
7 Scaling & Downhole Telemetry Loss with Wellhead Pressure Buildup \\
8 Hydrate (P) & Flow Restriction with Thermal Cooling and Telemetry Loss \\
9 Hydrate (S) & \textit{No convergence} (6 distinct physically-grounded names) \\
\bottomrule
\multicolumn{2}{l}{\textsuperscript{$\dagger$}\footnotesize Derived from word-frequency analysis of 51 unique names; no single name converged.}
\end{tabular}
\end{table}

\subsubsection{The ``telemetry loss'' meta-finding}

A single pattern dominates classes 2, 6, 7, and 8: the agent frames the novelty as a \textit{downhole telemetry dropout with a wellhead-side symptom}. This is an artifact of the 3W data---many anomaly onsets coincide with NaN or zeroed downhole sensors as gauges fail during the event. This is actionable feedback for the grouping layer: novelty clusters converging on ``telemetry loss'' should be split by their secondary symptom before being shown to the engineer.

\subsubsection{Class 9: richer naming, weaker detection}

Class~9 detection is the lowest (55\%, 6/11), but every flagged segment receives a distinct, physically-grounded name: \textit{Localized Upstream Restriction with Thermal Signature}, \textit{Inverse Pressure Divergence with Temperature Anomaly}, \textit{Downhole Gas Expansion Instability}, \textit{Total Sensor Signal Loss}, \textit{Thermal Inversion Restriction Event}, and \textit{Uncommanded Choke Opening}. The agent never maps missed class-9 segments onto the near-duplicate class~8, instead distributing them across class~2 (3 cases) and class~5 (2 cases) with sensor-reasonable justifications.

The practical consequence: \textbf{class-9-like novelty clusters may not consolidate under a single label} and should be surfaced as a family of distinct sub-events.

%% ============================================================
\section{Performance Summary}\label{sec:summary}
%% ============================================================

\begin{figure}[ht!]
\centering
\includegraphics[width=0.95\textwidth]{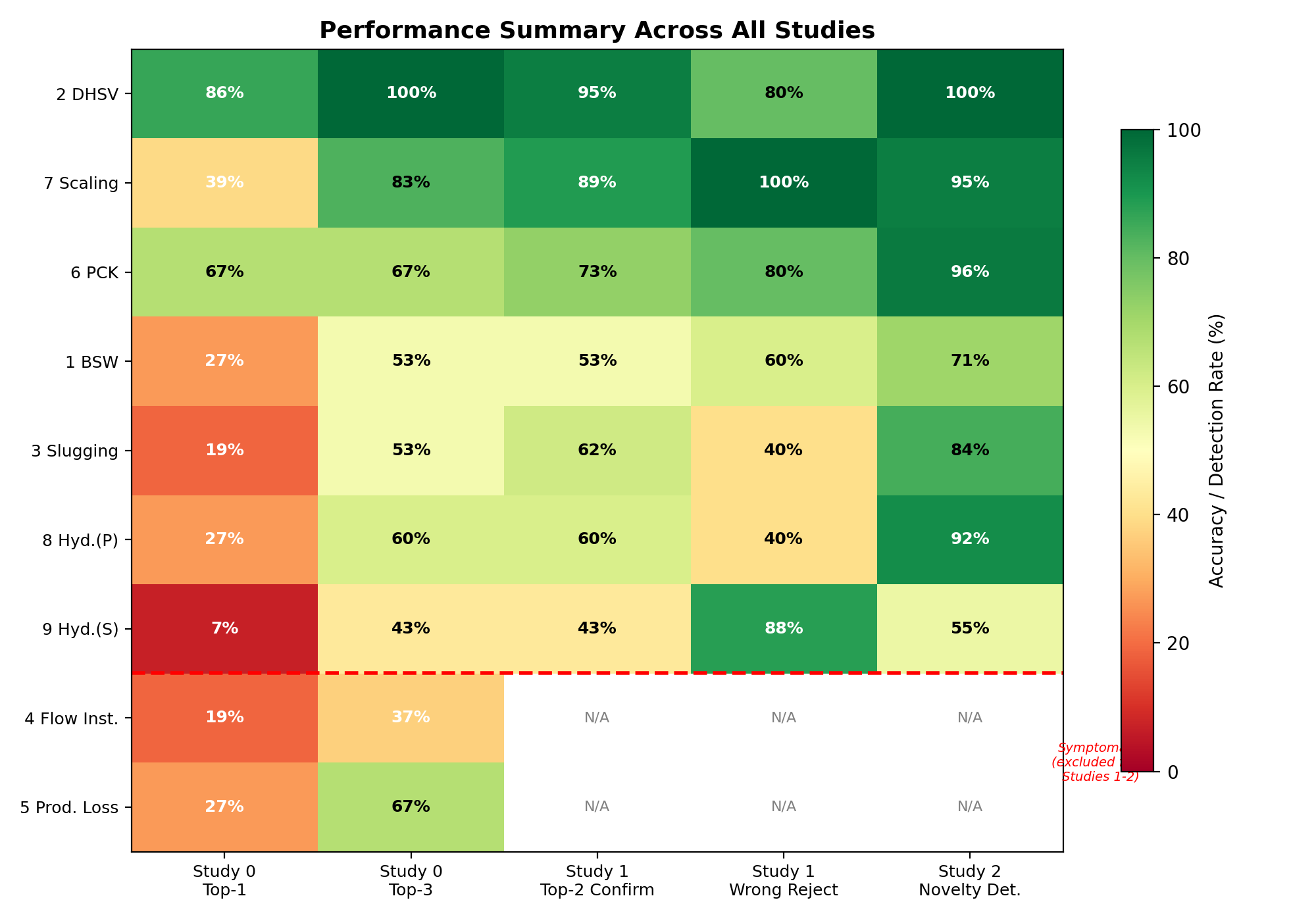}
\caption{Performance summary across all studies. Green = strong ($>$80\%), yellow = moderate (40--80\%), red = weak ($<$40\%). Classes below the dashed line are symptomatic and excluded from Studies 1--2.}
\label{fig:summary}
\end{figure}

Table~\ref{tab:summary} consolidates the key metrics.

\begin{table}[ht!]
\centering
\caption{Summary of key metrics across all studies.}
\label{tab:summary}
\begin{tabular}{lll}
\toprule
\textbf{Metric} & \textbf{Value} & \textbf{Interpretation} \\
\midrule
Study 0 top-3 (all 9 classes) & 63.9\% & Baseline, no upstream hint \\
Study 1 validates correct & 67.1\% & Correct proposals judged VALID \\
Study 1 rejects wrong & 73.7\% & Wrong proposals judged INVALID \\
Study 1 precision / F1 & 0.91 / 0.80 & Strong confirmation reliability \\
Study 2 novelty detection & 89.7\% & Flags held-out classes as novel \\
Study 2 name convergence & 6/7 classes & Stable per-cluster names \\
\bottomrule
\end{tabular}
\end{table}

%% ============================================================
\section{Discussion: A Companion to Published Methods}\label{sec:innovation}
%% ============================================================

The upstream OWL pipeline answers \textit{what happened}, \textit{where} (U-Net boundaries, 94\% IOU), and \textit{how confident} (classifier probability, Mahalanobis score). The LLM agent layer supplies the missing capabilities:

\begin{itemize}
    \item \textbf{Why does the model believe that?} Sensor-grounded justification citing specific metric deltas and cross-sensor relationships.
    \item \textbf{Could the upstream call be wrong?} Top-2 ranking + independent validity flag (73.7\% rejection of wrong proposals).
    \item \textbf{What should I do next?} Class-specific action recommendations keyed to cited evidence.
    \item \textbf{What is this novelty called?} Consolidated cluster names turning unlabeled clusters into engineer-readable labels.
    \item \textbf{When should I not trust the label?} Explicit flagging of symptomatic classes with prompt to re-rank.
\end{itemize}

Table~\ref{tab:division} shows how the division of labor plays out.

\begin{table}[ht!]
\centering
\caption{Division of labor between upstream OWL pipeline and LLM agent.}
\label{tab:division}
\begin{tabular}{lll}
\toprule
\textbf{Class} & \textbf{Upstream role} & \textbf{LLM agent role} \\
\midrule
2, 7 (clean) & Primary classifier & Confirm + justify + validate \\
1, 3, 6, 8 (moderate) & Primary, sometimes uncertain & Second opinion, name novelties \\
4, 5 (symptomatic) & Mahalanobis-rescue & Flag symptomatic, re-rank to top-3 \\
9 (heterogeneous) & Authoritative & Skeptic: confirm canonical only; name novelties \\
\bottomrule
\end{tabular}
\end{table}

%% ============================================================
\section{Limitations and Threats to Validity}\label{sec:limitations}
%% ============================================================

\begin{itemize}
    \item \textbf{Classes 4/5 ambiguity is physical.} No amount of reasoning over pressure and temperature metrics will separate symptoms from root causes. The deployment path is to re-rank to top-3 and treat the output as a differential diagnosis.
    \item \textbf{Class 9 cannot be reliably classified alone.} Study~0 top-1 is $7.1\%$ [1.3, 31.5] and Study~1 retains only $43\%$ of correct proposals. The upstream OWL classifier with spatial features remains authoritative.
    \item \textbf{Telemetry-loss confound.} The agent leans on downhole sensor dropouts when naming classes 2, 6, 7, and 8 (Section~\ref{sec:study2}). This is a 3W data artefact: many anomaly onsets coincide with NaN or zeroed downhole gauges. The reported $89.7\%$ overall detection rate therefore conflates physical anomaly characterisation with detection of telemetry interruptions. A focused re-evaluation restricted to segments with continuous downhole telemetry is left to future work, and the operational recommendation is to split novelty clusters by their secondary (non-telemetry) symptom before presenting them to engineers.
    \item \textbf{Class 1 has no natural name convergence.} Across 51 correctly-flagged novelties, all names are unique. A consolidated label (\textit{Wellhead Thermal-Pressure Surge}) was derived post-hoc from word-frequency analysis; the agent itself never converged organically.
    \item \textbf{Small per-class samples.} Several classes have $n \le 15$, yielding wide Wilson 95\% CIs (Tables~\ref{tab:study0},~\ref{tab:study2}). Differences between mid-performing classes are not statistically distinguishable. This study should be read as a feasibility evaluation, not a definitive ranking.
    \item \textbf{No baselines or ablations (future work).} The present work does not compare against (i) post-hoc explainability methods on the upstream classifier (e.g., SHAP, integrated gradients), (ii) alternative LLMs of different scale or family, or (iii) ablations of the prompt (metrics-only, profile-only, raw-window). These comparisons are required to attribute the agent's contribution to the LLM rather than to the structured-metrics pipeline that feeds it, and are planned as the immediate next step. We note, however, that the ceiling against which any such ablation must be measured is itself low: \citet{Lopes2025EAAI} and \citet{Lopes2026OTC} both document that classification and especially novelty detection on the 3W dataset under an open-world setting are far from trivial---binary classifiers remain overconfident on unseen events, class~5 (\textit{Rapid Productivity Loss}) defeats reconstruction-error detectors, and only a hybrid Mahalanobis latent-layer detector with deliberate engineering rescues novelty performance. A ``metrics-only'' or ``profile-only'' ablation should therefore be expected to leave several classes essentially undetectable, and we will report those expected failure modes alongside any positive findings.
    \item \textbf{Single inference run.} All reported numbers come from a single decoding pass per segment; LLM stochasticity is not characterised. Repeated sampling with majority voting and a temperature sweep are needed to bound run-to-run variance.
    \item \textbf{No proprietary data.} Evaluation used only the public 3W subset; the ICV Fault and other proprietary anomaly classes used in \citet{Lopes2025EAAI} are not included.
    \item \textbf{LLM grounding and hallucination.} The agent is general-purpose and not fine-tuned on oil well data. Although justifications cite sensor metrics provided in the prompt, the recommended-action and ``general knowledge'' fields are not independently validated and may contain plausible-sounding but incorrect domain claims. An expert audit of a stratified sample of justifications is left as future work.
\end{itemize}

%% ============================================================
\section{Conclusions}\label{sec:conclusions}
%% ============================================================

This work demonstrated that a general-purpose LLM (Qwen 3.5 MoE 397B), when provided with structured sensor metrics and data-driven anomaly profiles, can serve as an effective explainability and validation companion to an established Open-World Learning pipeline for oil well anomaly detection. The agent achieves 89.7\% novelty detection with physically-grounded naming, 91\% precision in validating upstream classifications, and generates sensor-specific justifications that bridge the gap between numerical scores and operator understanding.

The key contribution is not classification accuracy---the upstream pipeline outperforms the LLM on that dimension---but rather the three capabilities the upstream pipeline cannot provide: natural-language justification of decisions, independent validation with rejection of implausible labels, and consolidated naming of novelty clusters. These capabilities directly address the scalability limitation identified in \citet{Lopes2025EAAI} regarding operator-driven validation, and complement the improved novelty detection introduced by \citet{Lopes2026OTC}: the LLM agent provides preliminary naming and characterization of novel anomaly clusters flagged by the Mahalanobis latent-layer detector, reducing the burden on human experts while maintaining interpretability.

Future work will explore fine-tuning domain-specific LLMs on oil well sensor data, extending the evaluation to proprietary anomaly types (ICV Fault, Valve Operation), and integrating the agent layer into a real-time monitoring interface.

%% ============================================================
\section*{Acknowledgments}
%% ============================================================

The authors thank PETROBRAS for supporting this research and for providing access to data and domain expertise.

%% ============================================================
\section*{CRediT authorship contribution statement}
%% ============================================================

\textbf{Lucas Gouveia Omena Lopes}: Conceptualization, Data curation, Formal analysis, Methodology, Software, Validation, Writing -- original draft, Writing -- review \& editing.
\textbf{Thales Miranda de Almeida Vieira}: Formal analysis, Methodology, Supervision, Writing -- review \& editing.
\textbf{Pedro Esteves Aranha}: Formal analysis, Methodology, Supervision, Writing -- review \& editing.
\textbf{Eduardo Toledo de Lima Junior}: Formal analysis, Investigation, Methodology, Supervision, Writing -- original draft, Writing -- review \& editing.
\textbf{William Wagner Matos Lira}: Conceptualization, Formal analysis, Investigation, Supervision, Writing -- review \& editing.

\section*{Declaration of competing interest}
The authors have no relevant financial or non-financial interests to disclose.

\section*{Data availability}
The 3W dataset is publicly available \citep{VARGAS2019}. The code that generates the structured metrics, builds the class profiles, runs the agent against the NVIDIA NIM endpoint, and produces all tables and figures in this paper, together with the raw per-segment JSON outputs of all three studies, will be released in a public GitHub repository with an archived Zenodo DOI upon acceptance.

%% ============================================================
\appendix
%% ============================================================

\section{Sample Knowledge Profile}\label{app:profile}

The following is a condensed version of the structured profile provided to the LLM for Class~2 (Spurious Closure of DHSV). Each of the nine classes has an analogous profile built from training data.

\begin{verbatim}
# Anomaly Profile: Spurious Closure of DHSV (Class 2)
## Built from 38 files (38 segments)

### P-PDG
  Behavior: step_change | Direction: increase
  Magnitude: large (>20%) | Rate: gradual (>60s)
  Typical delta: mean=0.903 [0.856 .. 0.949]
  Macro: highly consistent step_change (rising)
  Local: plateau, gradual onset, no recovery

### P-TPT
  Behavior: step_change | Direction: decrease
  Magnitude: large (>20%) | Rate: gradual (>60s)
  Typical delta: mean=-0.729 [-0.863 .. -0.596]
  Macro: highly consistent step_change (dropping)

### T-PDG
  Behavior: step_change | Direction: decrease
  Magnitude: large (>20%) | Rate: gradual (>60s)
  Typical delta: mean=-0.414 [-0.623 .. -0.195]
  Macro: variable (frequent no_data)

### T-TPT
  Behavior: step_change | Direction: decrease
  Magnitude: large (>20%) | Rate: gradual (>60s)
  Typical delta: mean=-0.835 [-0.921 .. -0.750]

## Distinguishing Features
- P-PDG strongest rise (large, gradual)
- T-TPT strongest drop (large, gradual)
- P-PDG and P-TPT DIVERGE in direction

## General Knowledge
  Causes: control line failure, hydraulic leak
  Severity: critical
  Related: Hydrate in Production Line, Flow Instab.
\end{verbatim}

\section{Metric Computation Details}\label{app:metrics}

Given preprocessed baseline $\mathbf{b}$ and event $\mathbf{e}$ windows:

\begin{align}
\delta &= \bar{e} - \bar{b} \\
\text{pct} &= \frac{\delta}{|\bar{b}|} \times 100 \quad (\text{guarded: } |\bar{b}| > 10^{-12}) \\
z &= \frac{|\delta|}{\sigma_b} \quad (\sigma_b = \max(\text{std}(\mathbf{b}), 10^{-6})) \\
\text{slope} &= \arg\min_{a,c} \sum_t (e_t - a \cdot t - c)^2
\end{align}

Profile aggregation uses two-pass IQR filtering: first with $k=3.0$, then $k=1.5$. Points outside $[Q_1 - k \cdot \text{IQR},\; Q_3 + k \cdot \text{IQR}]$ are removed before computing summary statistics. This guards against near-zero-baseline explosions while preserving the core distribution.

\bibliographystyle{elsarticle-harv}
\bibliography{references}

\end{document}